\documentclass{article}
\usepackage{picins}
\usepackage{spconf,amsmath,graphicx}
\usepackage[table,xcdraw]{xcolor}

\usepackage{enumitem}
\setlist{nosep, leftmargin=14pt}
\usepackage{hyperref}

\usepackage{mwe} 
\usepackage{multirow}

\title{COLON: The largest \textbf{CO}lonoscopy \textbf{LON}g sequence public database}
\name{Lina Ruiz $^{\star}$, Franklin Sierra-Jerez $^{\star}$, Jair Ruiz $^{\dagger}$, Fabio Martínez $^{\star}$}

\address{Biomedical Imaging, Vision, and Learning Laboratory (BIVL$^{2}$ab),\\ Universidad Industrial de Santander (UIS), Colombia
$^{\star}$ \\ Instituto de Gastroenterología y Hepatología del Oriente Colombiano (IGHO S.A.S.), Colombia $^{\dagger}$
\thanks{$^{1}$ Biomedical Imaging, Vision and Learning Laboratory (BIVL$^2$ab). Universidad Industrial de Santander (UIS), Colombia \newline {famarcar@saber.uis.edu.co}}
\thanks{$^{2}$Instituto de gastroenterología y hepatología del oriente IGHO S.A.S}}%

\begin{document}
\maketitle

\begin{abstract}

Colorectal cancer is the third most aggressive cancer worldwide. Polyps, as the main biomarker of the disease, are detected, localized, and characterized through colonoscopy procedures. Nonetheless, during the examination, up to 25\% of polyps are missed, because of challenging conditions (camera movements, lighting changes), and the close similarity of polyps and intestinal folds. Besides, there is a remarked subjectivity and expert dependency to observe and detect abnormal regions along the intestinal tract. Currently, publicly available polyp datasets have allowed significant advances in computational strategies dedicated to characterizing non-parametric polyp shapes. These computational strategies have achieved remarkable scores of up to 90\% in segmentation tasks. Nonetheless, these strategies operate on cropped and expert-selected frames that always observe polyps. In consequence, these computational approximations are far from clinical scenarios and real applications, where colonoscopies are redundant on intestinal background with high textural variability. In fact, the polyps typically represent less than 1\% of total observations in a complete colonoscopy record. This work introduces \textbf{COLON:} the largest \textbf{CO}lonoscopy \textbf{LON}g sequence dataset with around of 30 thousand polyp labeled frames and 400 thousand background frames. The dataset was collected from a total of 30 complete colonoscopies with polyps at different stages, variations in preparation procedures, and some cases the observation of surgical instrumentation. Additionally, 10 full intestinal background video control colonoscopies were integrated in order to achieve a robust polyp-background frame differentiation. The \textbf{COLON} dataset is open to the scientific community to bring new scenarios to propose computational tools dedicated to polyp detection and segmentation over long sequences, being closer to real colonoscopy scenarios. 
\end{abstract}


\section{Introduction}

Colorectal cancer (CRC) is the third most common cancer and the second most deadly cancer worldwide \cite{ferlay2021cancer}. Polyps, as the main CRC biomarker, are abnormal masses that grow in the intestinal tract, whose detection and shape characterization constitute a key factor in the patient's survival rate. These polyps are typically observed from colonoscopy procedures, but their detection and respective characterization require exhaustive observations, taking an average of 20 to 30 minutes. Besides, during colonoscopy procedures, the polyp may present high appearance variability due to constant illumination changes, artifacts in the intestinal tract, and abrupt camera movements. Several studies have reported that during the clinical procedure between 6-25\% of polyps are missed \cite{angermann2016active}. This critical issue affects the early diagnosis impacting directly the survival rate, \textit{i.e.,} patients at the fourth stage have an 8\% of survival probability \cite{gastroenterologa}.

Computer aid diagnosis systems have emerged as clinical supporting tools contributing to polyp characterization. Recently, the state-of-the-art has reported remarkable advances in several tasks related to polyp detection, segmentation, and classification \cite{fan2020pranet, qadir2021toward, zachariah2020prediction}. Such advances have been developed thanks to the availability of public datasets with polyp observations from isolated frames and short video sequences of diverse colonoscopy procedures \cite{tajbakhsh2015automated, angermann2017towards, bernal2018polyp}. In fact, several strategies report polyp segmentation scores up to 90\%, which looks to overcome the problem of polyp characterization. Nonetheless, typical colonoscopy procedures are very long and exhaustive, taking more than 20 minutes per procedure, with around 16 thousand frames. From such colonoscopies, the polyp observations appear in less than 1\% of the frames. Beyond the textural and light variations, many of these procedures have additional challenges such as poor patient preparation and intestinal folds that may be confused with polyp masses. So, far from resolving the polyp detection and characterization problem, the community needs to develop new strategies to operate in scenarios closer to clinical procedures that consider many of the challenges during colonoscopies.

\begin{figure*}[ht]
    \centering
    \includegraphics[width=0.9\textwidth]{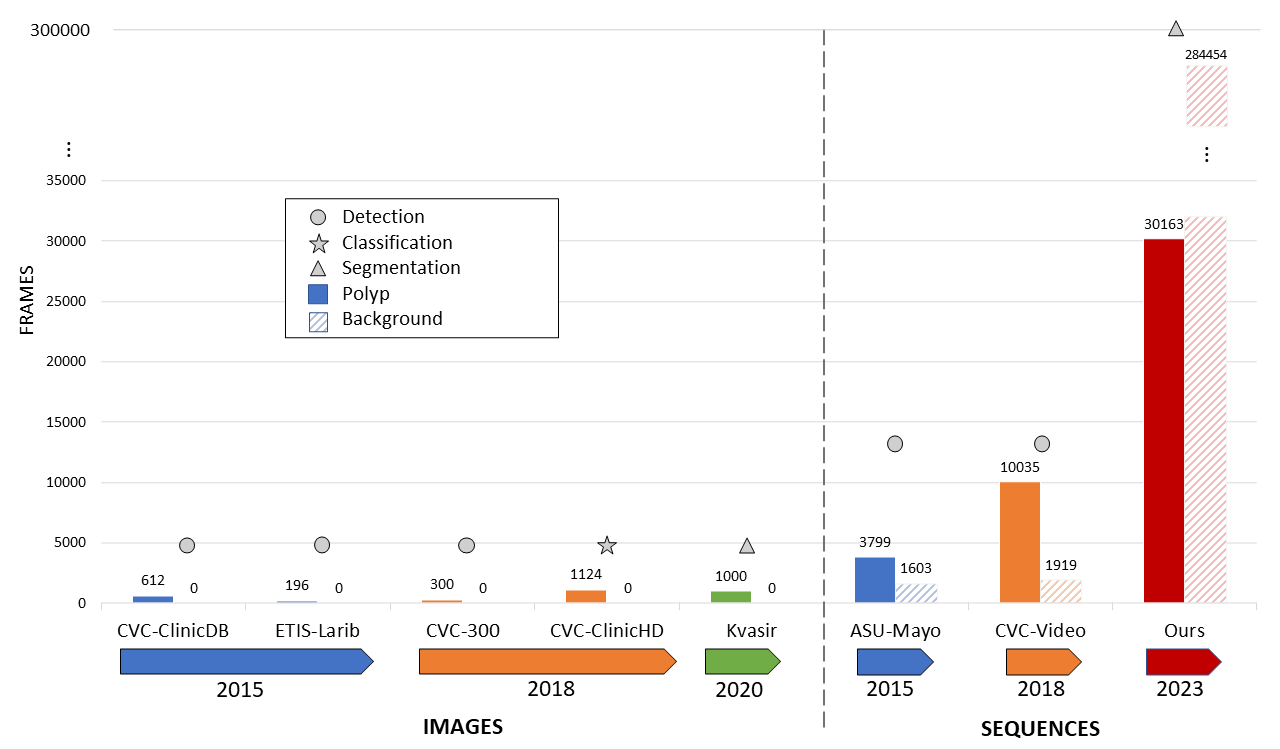}
    \caption{Comparison between public datasets (CVC-ClinicDB \cite{bernal2015wm}, ETIS-Larib \cite{silva2014toward}, CVC-300 \cite{bernal2012towards}, CVC-ClinicHD  \cite{sanchez2019computer, bernal2019gtcreator}, Kvasir \cite{Pogorelov:2017}, ASU-Mayo \cite{tajbakhsh2015automated}, CVC-Video \cite{angermann2017towards, bernal2018polyp}) available since 2015 and our proposed dataset for 2023 associated to \textbf{COLON} challenge.}
    \label{public_datasets}
\end{figure*}

This work introduces \textbf{COLON:} the largest \textbf{CO}lonos\-copy \textbf{LON}g sequence dataset with more than 30 thousand labeled polyp frames and 400 thousand intestinal fold background frames. These video sequences were collected from 30 colonoscopies with polyps at different stages, and procedures with remarked visual variability. A total of 30 videos labeled by an expert gastroenterologist were released for training (20 publicly available) and testing (10 as a private set), respectively.  In addition, 10 non-polyp sequences (5 for split) were added to accomplish the proper polyp and non-polyp frame differentiation. The labeled sequences allow the exploration of computational capabilities to detect and segment polyps over long video sequences. This work also presents an analysis of current state-of-the-art techniques, running over \textbf{COLON} dataset, and discusses new guidelines and perspectives of computational approaches to operate in real scenarios.

\section{Related works and datasets}


Nowadays, there exist multiple open datasets that include polyp observations from colonoscopies, allowing the design, implementation, and adjustment of computational representations to support several polyp characterization tasks. For instance, the first available datasets were the \textit{CVC-Clinic DB}\footnote{\url{https://polyp.grand-challenge.org/CVCClinicDB/}} (612 polyp frames with respective masks from 31 colonoscopies) and the \textit{ETIS-Larib} \footnote{\url{https://polyp.grand-challenge.org/Databases/}} (196 polyp frames with respective ground truth masks from 34 colonoscopies). These datasets were designed for polyp detection tasks, allowing to development of methods dedicated to model polyp shape features \cite{bernal2015wm, silva2014toward}. Around this data, several analyses were proposed to capture non-linear polyp-background features from receptive field blocks (dilated convolutions enhance local representations) \cite{fan2020pranet, huang2021hardnet}.

Also, the \textit{CVC-300} database \footnote{\url{http://adas.cvc.uab.es/endoscene}} was publicly available to bring images of polyps with additional morphological variability (flat and peduncular). This database consists of 300 annotated polyp images extracted from 15 different videos \cite{bernal2012towards}. Besides, the \textit{CVC-ClinicHD} dataset (composed by more than 1100 HD images showing a polyp) was an effort to bring colonoscopy data with major descriptions about polyp malignancy classes \footnote{\url{https://giana.grand-challenge.org/Home/}}. These improved datasets open new polyp-related challenges not only related to size, and morphology but also surface patterns to address biopsy estimations \cite{mareth2022endoscopic, mori2020cost}. More recently, the \textit{Kvasir-SEG} database, composed of 1450 annotated polyp frames was published in 2020 for polyp segmentation task \footnote{\url{https://datasets.simula.no/kvasir-seg/}}. This open-access dataset brings new computer-aid diagnosis systems around the segmentation, detection, and localization of polyps \cite{jha2019resunet++}.

Short colonoscopy sequences were also added to propose challenges in close realistic scenarios that include intestinal background, i.e, frames without polyps. For instance, the \textit{ASU-Mayo} database, published in the Automatic Polyp Detection challenge in 2015 \footnote{\url{https://polyp.grand-challenge.org/AsuMayo/}}, contains for training purposes, 20 short colonoscopy sequences (one minute on average) balanced between records with polyp and only intestinal folds. Regarding the test split, this dataset has 18 videos without public ground truth annotations. In the same line, the \textit{CVC-Video}, published in the Endoscopic Vision Challenge in 2018 \footnote{\url{https://giana.grand-challenge.org/}}, is composed of 18 short colonoscopy sequences for training and test (one minute on average) with polyps/non-polyps and the corresponding binary ground truth annotation. Around these more realistic scenarios, several computational techniques that include deep representations and state-of-the-art techniques, have been proposed to address the polyp characterization \cite{fan2020pranet, huang2021hardnet}.

Despite the availability of colonoscopy data and current computational approaches, there is a main drawback to transferring these solutions to real scenarios. Essentially, the approaches have been designed over restricted and cropped scenarios, which is a potential limitation to include challenges typically observed in colonoscopies. In fact, in the clinical routine, the colonoscopies easily exceed thousands of frames meaning more challenging visual variations. The polyp characterization problem remains open, and the community needs to develop new computer-aid diagnosis strategies that will be able to operate in scenarios closer to real practice.

\section{\textbf{COLON:} the largest \textbf{CO}lonoscopy \textbf{LON}g sequence dataset}

This work presents the \textbf{COLON} dataset, which collects the major number of polyp and non-polyp frame observations. Regarding the current open dataset in the state-of-the-art, this dataset includes recovered samples extracted from 30 colonoscopy video sequences with polyps, from typical procedures with a huge amount of intestinal background frames together with polyp findings with high visual and shape variability. Figure \ref{public_datasets} reports a comparison between the proposed dataset and the baseline datasets introduced in the state-of-the-art. As observed, the proposed dataset exceeds the amount of available data, and the labeled information, introducing the task of polyp segmentation task in long sequences and allowing approximate strategies to real scenarios.

\subsection{Data}

The COLON dataset \footnote{\textbf{COLON} dataset was elaborated in collaboration with \textit{Instituto de Gastroenterología y Hepatología del Oriente - IGHO S.A.S.} and the Biomedical Imaging, Vision, and Learning Laboratory - BIVL$^2$ab from \textit{Universidad Industrial de Santander} in Bucaramanga, Colombia.} includes colonoscopies recorded with Olympus Evis Excera III 190 colonoscopy, with a spatial resolution of $480\times720$ and a \-temporal resolution of $30$ \textit{fps}. Each video has an average of 16 thousand frames with unbalanced observations since most of the frames present only the intestinal tract. The dataset contains 30 long sequences with polyps and 10 long sequences without polyps and at least three sequences have two polyps. The polyps are highly variable regarding size, NICE classification, morphology (sessile or pedunculated), and biopsy results (adenoma or hyperplastic). An expert gastroenterologist (with more than 13 years of experience) labeled these findings. In addition, demographic variables such as sex and age ($[35-95], \mu = 65, \sigma = 13.6 $) were also included in this study. 

\begin{figure}[ht]
    \centering
    \includegraphics[width=0.48\textwidth]{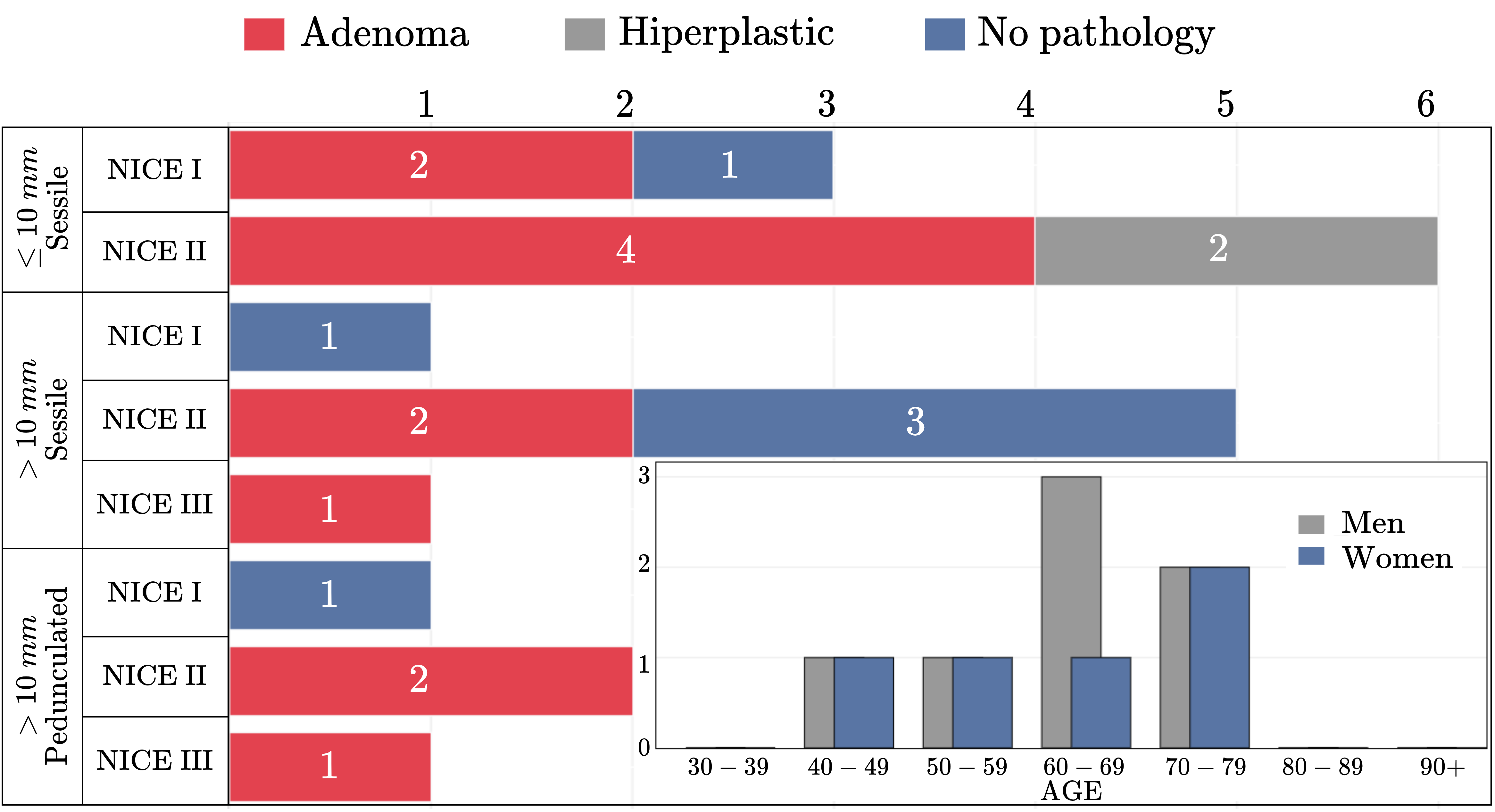}
    \caption{Polyp description according to the size, the morpho\-logy (sessile or pedunculated), NICE classification, and the biopsy result. The bottom figure shows the demographic information.}
    \label{information_dataset}
\end{figure}

Figure \ref{information_dataset} illustrates data distribution of \textbf{COLON} dataset based on demographic variables such as age and gender. As observed, the majority of patients are older than 50 years, with a gender distribution, except for the interval of 60-69 years (two additional samples for men). The dataset was also analyzed in terms of polyp malignancy, following the NICE protocol, and also regarding the histopathological classification: adenoma, hyperplastic, and non-pathological cases. This analysis is valuable for quantifying the capabilities of strategies to correlate detection with malignancy and stratifying polyps. Furthermore, the \textbf{COLON} dataset has been categorized and quantified based on polyp size, a main macroscopic feature associated with malignancy. It should be noted that many of the recorded sequences have polyps smaller than 10 \textit{mm}, which results in a challenge for strategies, but with major value in supporting detection efforts.

\begin{figure*}[ht]
    \centering
    \includegraphics[width=0.9\textwidth]{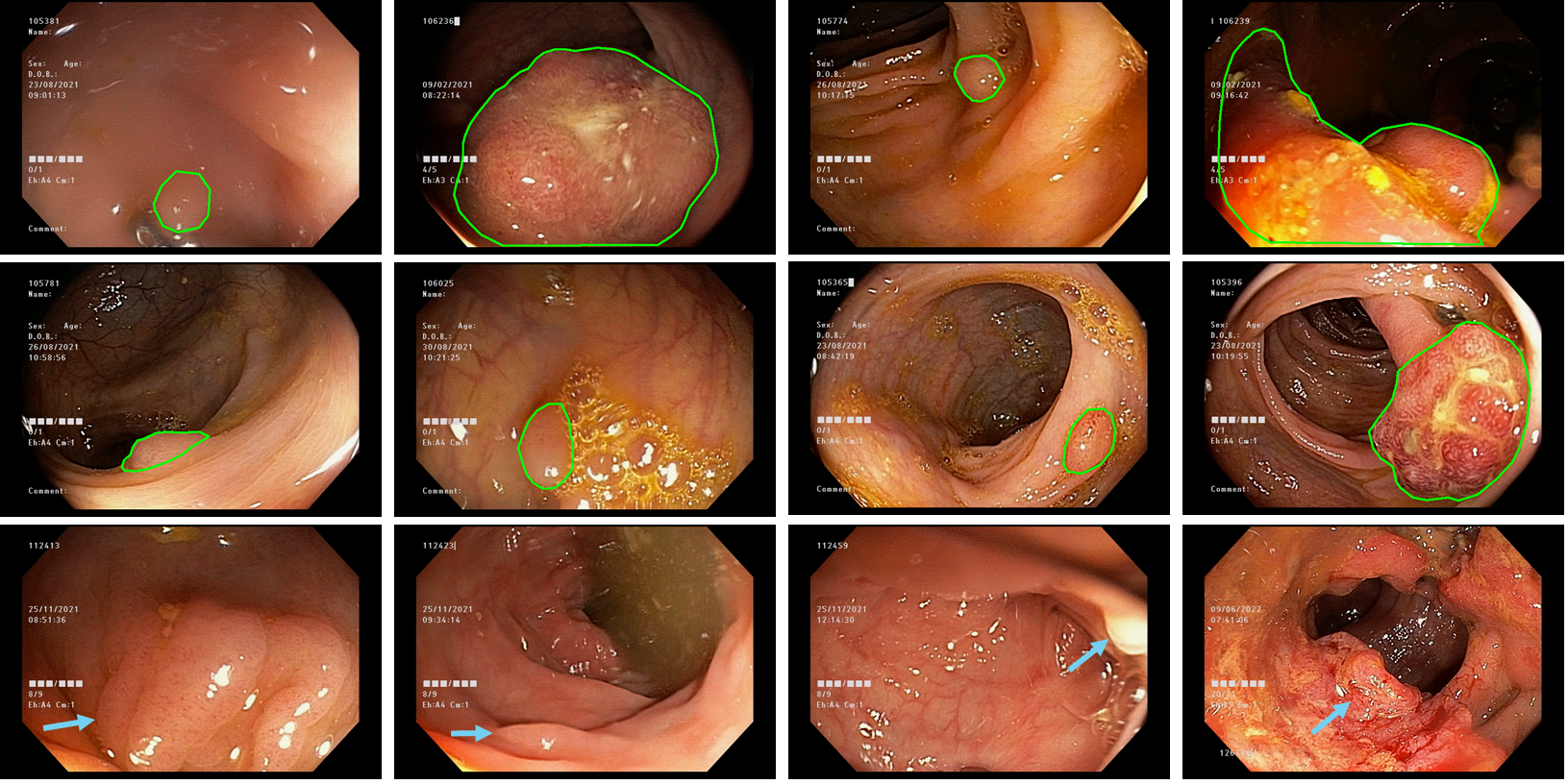}
    \caption{Frames extracted from the captured colonoscopy sequences. The first two rows contain polyps with their respective marking (green contour). The bottom row shows typical intestinal regions prone to be misidentified as polyps due to their similar polyp patterns (blue arrows).}
    \label{data_polyps}
\end{figure*}

Figure \ref{data_polyps} shows visual representations of various polyps. highlighting their morphological, size, and texture variability. Additionally, it showcases the impact of water bubbles and intestinal residues in the extracted samples, adding complexity to the detection task. The green contours in the images represent the boundaries of polyps taken from the expert annotation. In the last row, the blue arrows draw attention to some structures that exhibit patterns resembling polyps, such as specular reflections or mucous. These observations complement the dataset analysis mentioned earlier, emphasizing the challenges associated with accurate polyp recognition.

\subsection{Polyp segmentation and localization tasks}

The \textbf{COLON} dataset was split into training and test subsets in order to establish a general methodology to compare computational strategies. Such partition is considered a split from colonoscopies to avoid overfitting of textural intestinal tract patterns. Both sets include colonoscopy videos without any polyp findings. Table \ref{data_distribution} depicts an overview of the data distribution within the training (public) and the test (private) sets. The training sequences were labeled at intervals of every 10 frames, while dense labeling, involving annotation for every frame of the sequence, was applied to the test set. For this set of sequences, we propose the localization and segmentation tasks, which will be validated according to the long sequences. The tasks are explained as follows:\\

\begin{table}[ht]
\begin{center}
\begin{tabular}{|c|c|c|}
    \hline
    \multicolumn{3}{|c|}{\textbf{Train}}\\ \hline
    \multicolumn{2}{|c|}{\multirow{2}{*}{20 videos with polyp}} &  5 videos \\ 
    \multicolumn{2}{|c|}{} & without polyp \\ \hline
    \textit{Polyp}  & \textit{Background}  & \textit{Background} \\  
   $> 1800 $ frames &  $> 128000 $ frames & $> 48000 $ frames  \\ \hline
\end{tabular}
\end{center}

\begin{center}
\begin{tabular}{|c|c|c|}
    \hline
    \multicolumn{3}{|c|}{\textbf{Test}}\\ \hline
    \multicolumn{2}{|c|}{\multirow{2}{*}{10 videos with polyp}} &  5 videos \\ 
    \multicolumn{2}{|c|}{} & without polyp \\ \hline
    \textit{Polyp}  & \textit{Background}  & \textit{Background} \\  
   $> 28000 $ frames &  $> 150000 $ frames & $> 48000 $ frames  \\ \hline
\end{tabular}
\end{center}
\caption{The distribution between train and test set of the dataset associated with \textbf{COLON} dataset.}
\label{data_distribution}
\end{table}

    \begin{itemize}
    
    \item \textbf{Polyp segmentation.} This task has been widely addressed in recent years. The primary goal is to estimate the polyp-shape feature (correlated with the malignancy degree). Nonetheless, this task remains challenging due to the high visual variability observed in real colonoscopy sequences. This visual variability is not only influenced by the polyp's morphology but also by the clinical inherent challenges like strong camera movements and poor light conditions, among others. In this sense, in real colonoscopy sequences, the polyp characterization in a frame-by-frame strategy remains unpredictable considering that even neighboring frames may contain more intestinal background than polyp information or vice versa. For the COLON dataset we decided to codify an overlapping segmentation metric, but also including the frames without polyp observations. Then, we can recover polyp shape estimations $\hat{S}$, and estimate regions that easily are confused with polyp sample (blue arrows in figure \ref{data_polyps}).  In such cases, two scores are computed along each sequence,  for frames with polyp ($S_{wp}$), and frames with only background ($S_{ob}$), as: 

\begin{equation}
    S_t = \left\{\begin{matrix}
\text{With Polyp ($S_{wp}$)} & \to  & S_{wp} =\frac{S_i\cap\hat{S_i} }{S_i\cup \hat{S_i} } \\ 
                  &      &                                                      \\
\text{Only Back ($S_{ob}$)}   & \to &  S_{ob} =\frac{WH -\text{Positive pixels} }{WH}
\end{matrix}\right.
\label{eq_seg}
\end{equation}

From equation \ref{eq_seg} is then computed the segmentation scores for frames with ($S_{wp}$) and without ($S_{ob}$) polyps. Then, a linear combination of both scores allows to measure the impact of strategies in long sequences, as: 
$$S = \alpha (\mu(S_{wp})) + (1-\alpha)(\mu(S_{ob}))$$.

\item \textbf{Polyp localization.} Despite state-of-the-art works have traditionally approached the polyp segmentation task, the polyp localization task could not be typically addressed due to the scarce public data availability that offers the typical variability in the polyp-background information. In this sense, a True Positive (TP) prediction is considered when the IoU score is greater than a defined threshold ($\tau_{wp} \geq 0.5$), otherwise, it is categorized as a False Negative (FN). Furthermore, in the more common scenario where the frame captures the intestinal background, if the model's prediction ($S_{ob}$) is a black mask, then the predicted frame is classified as a True Negative (TN). However, if any frame region is misidentified as a polyp, then the frame is labeled as a False Positive (FP). Therefore, the polyp localization task is defined and assessed using the following metrics:
\end{itemize}

\begin{align*}
    \text{Specificity} = \frac{TN}{TN+FP} && \text{Precision} = \frac{TP}{TP+FP} \\
\end{align*}

\section{Baseline and evaluation}

As a baseline, three state-of-the-art strategies with remarkable results over traditional public datasets (report scores up to 80\%) were validated over the dataset associated with \textbf{COLON} challenge (\cite{fan2020pranet, huang2021hardnet, ruiz2022weakly}). Table \ref{baseline_seg} shows the achieved performance by each analyzed strategy, with variations in the significance of frames containing polyps ($S_{wp}$) and those without polyps ($S_{ob}$) through the $\alpha$ factor. Here, the IoU represents the average of the metric $S_t$ calculated for each frame. As expected, when exclusively frames with polyps are considered ($\alpha=1$), these methods exhibit lower scores in practical clinical scenarios, with a maximum of \textbf{69.4\%}. However, as the importance of polyp frames decreases, the percentage of IoU increases. This behaviour is evident for all strategies, which obtain around 80\% IoU with an $\alpha=0.3$.

\begin{table}[ht]
\begin{center}
\label{baseline_seg}
\begin{tabular}{|c|c|c|c|c|}
\hline
\multirow{2}{*}{\textbf{Method}} & \multicolumn{4}{|c|}{\textbf{IoU (\%)}} \\ \cline{2-5}
 & \textbf{\begin{tabular}[c]{@{}l@{}} $\alpha$ =1.0\end{tabular}} & \textbf{\begin{tabular}[c]{@{}l@{}} $\alpha$=0.7\end{tabular}} & \textbf{\begin{tabular}[c]{@{}l@{}} $\alpha$=0.5\end{tabular}} & \textbf{\begin{tabular}[c]{@{}l@{}} $\alpha$=0.3\end{tabular}} \\ \hline
Fan, D \cite{fan2020pranet} & \textbf{69,4}  & \textbf{75,2} & \textbf{79,1}  & 83,0 \\ \hline
Huang, C. H. \cite{huang2021hardnet} & 67,8 &  74,4 & 78,8 & 83,2  \\ \hline
Ruiz, L \cite{ruiz2022weakly} & 58,7 &  70,9 & \textbf{79,1} & \textbf{87,4} \\ \hline
\end{tabular}
\end{center}
\end{table}

Table \ref{baseline_localization} depicts the achieved results for the three considered strategies concerning the polyp localization task. As anticipated, a substantial quantity of background frames leads to a heightened occurrence of false positives.  This fact is a dramatic point since false positives may result in false alarms which delay the colonoscopy procedure, or even worse may result in mistakes in surgical procedures. To evaluate this scenario, for each frame, was admitted a certain region with positive detection, which may be deleted and avoid the false positive alarm. In this case, two region sizes were considered: background regions larger than 70\% ($\tau_{ob} = 0.70 $), and 95\% ($\tau_{ob} = 0.95$). It should be noted that regions with false positive regions of around 30\%  ($\tau_{ob} = 0.70$) are really big regions because small polyps (ranging from 5 mm to 40 mm) may have this size. However, to report results of computational results with different scopes we decided to cover these two thresholds.

\begin{table}[ht]
\begin{center}
    
    \begin{tabular}{|c|c|c|}
    \hline
    \multirow{2}{*}{\textbf{Method}} & \multicolumn{2}{|c|}{\textbf{$\tau_{ob} = 0.70$}} \\ \cline{2-3}
    &  \textbf{Precision (\%)} & \textbf{Specificity (\%)} \\ \hline
    Fan, D \cite{fan2020pranet} & 46,5 & 82,3     \\ 
    Huang, C.\cite{huang2021hardnet} & 45,4 & 82,1    \\ 
    Ruiz, L \cite{ruiz2022weakly}  & 98,7 & 99,8     \\  \hline
    & \multicolumn{2}{|c|}{\textbf{$\tau_{ob} = 0.95$}} \\ \hline
    Fan, D \cite{fan2020pranet} & 25,7 & 55,7     \\ 
    Huang, C.\cite{huang2021hardnet} & 29,2 & 64,1     \\ 
    Ruiz, L \cite{ruiz2022weakly} & 86,9 & 97,9 \\  \hline
    \end{tabular}
    \caption{Validation of the state-of-the-art architectures in the dataset associated to \textbf{COLON} challenge. These results are associated with the localization task.}
    \label{baseline_localization}
\end{center}
\end{table}

The results reveal a general decrease in performance across the three strategies as the threshold increases. In particular, architectures like Huang \textit{et al} and Fan \textit{et al} exhibit a notable prevalence of false positives. For instance, Fan reports a precision of 25,7\% and a specificity of 55,7\%, inferring that the recognition of frames with background is challenging. Additionally, this behavior can be attributed to the architectures being typically trained in constrained scenarios where the polyp is the predominant feature. Nevertheless, Ruiz achieved a precision of 86,9\% and a specificity of 97,9\% in the more challenging scenario ($\tau_{ob} = 0.95$) due to the weakly supervised strategy used in the training phase to learn the main polyp features, but also, characterize the background (intestinal tract). Notably, Huang reports a precision of 87,3\% in CVC-Video sequences while in the ASU-Mayo Clinic dataset, it was 83,4\%. However, in the dataset associated with \textbf{COLON} challenge the precision decreased to 29,2\%.

In this work, we built an official web site \footnote{\url{https://bivl2ab.uis.edu.co/challenges/colon}} to explain details of dataset and bring the possibility to scientific community to test models using \textbf{COLON} database. The participants should submit a docker file with the dependencies used, the model weights (file h5 for instance), and a script with the specifications for running the model. Each participant will count with a profile to track scores of each submission and the performance of their model. 

\section{Discussion and remarks}

This work presented a novel \textbf{COLON} dataset with long colonoscopy sequences. The main objective of this dataset is to naturally capture the colonoscopy procedure, obtaining sequences that last approximately 15 minutes. These videos involve strong camera movements and a large amount of background information (intestinal tract without polyp). Contrary to public datasets, \textbf{COLON} allows analyzing more realistic clinical environments due to the high polyp variability in terms of morphology, and the different scenarios (poor conditions and surgical instruments intervention) included in the capture protocol.

The achieved results from baseline computational strategies demonstrated the current challenges to transfer such technology in real scenarios.  In fact, such strategies have remarked limitations in the \textbf{COLON} dataset to operate into scenarios with redundant background. Architectures such as HarDNet obtained on public datasets a mIoU of around 80\% but in \textbf{COLON} dataset this percentage decrease to 67\%. Furthermore, some methods only learn to identify the polyp region and attempt to generate predictions for all presented frames. This behavior is particularly evident in architectures like HarDNet and PraNet, which exhibit a precision lower than 30\%. 

In this work was reported a public database with a long colonoscopy sequence closed the computational strategies to real scenarios. This project has available an official website and a platform to validate new proposals. Researchers are invited to adjust arquitectures with colon available data, and then to submit solutions to the platform, which be coordinated to publish the score table of the best participants. 

\bibliographystyle{IEEEbib}
\bibliography{refs}

\end{document}